\documentclass[letterpaper, 10 pt, conference]{ieeeconf}

\IEEEoverridecommandlockouts
\usepackage{amsmath}
\usepackage{amssymb}
\usepackage{graphicx}
\usepackage{booktabs}
\usepackage{multirow}
\usepackage[colorlinks=true, allcolors=blue]{hyperref}

\newcommand{\catlazy}{\textsc{Cat-Lazy}}
\newcommand{\catactive}{\textsc{Cat-Active}}
\newcommand{\reslazy}{\textsc{Res-Lazy}}
\newcommand{\resactive}{\textsc{Res-Active}}

\title{\LARGE \bf
Wiring Matters: Injection Topology and Initialization\\ of Affordance Heads in Vision-Language-Action Policies
}

\author{Zijian An$^{1}$, Linhan Wang$^{2}$, Jiayan Wang$^{3}$, Shijie Geng$^{4}$, Ran Yang$^{2}$, Yiming Feng$^{2}$, and Lifeng Zhou$^{1*}$
\thanks{$^{1}$Drexel University, Philadelphia, PA 19104, USA
        {\tt\small \{za382, lz457\}@drexel.edu}}%
\thanks{$^{2}$Virginia Tech, Virginia Seafood Agricultural Research and Extension Center, Hampton, VA 23669, USA
        {\tt\small \{ryang17, yimingfeng\}@vt.edu}}
\thanks{$^{3}$TODO: Jiayan Wang's affiliation and email}%
\thanks{$^{4}$Amazon Store Foundation AI (SFAI), New York, NY 10018, USA
        {\tt\small jeykigung@gmail.com}. This work is independent of and outside of the author's work at Amazon.}%
\thanks{$^{*}$Corresponding author.}%
}

\begin{document}

\maketitle
\thispagestyle{empty}
\pagestyle{empty}

\begin{abstract}
Dense affordance supervision is an appealing auxiliary signal for vision-language-action (VLA) policies, yet naively co-training an
  affordance head can severely damage instruction following. We present a controlled study of \emph{how} such a head should be wired into
  a modern VLA on the LIBERO benchmark. Our recipe reads the backbone through a stop-gradient and re-injects an intermediate head feature
  into the action expert via a learned bridge. The stop-gradient is a precondition: letting affordance gradients reach the backbone drops
  the policy below the headless base ($85.5\%$ vs.\ $93.1\%$). With the backbone protected, a same-budget $2{\times}2$ ablation over
  injection topology (concatenation vs.\ residual) and bridge initialization (zero vs.\ random) shows initialization is the dominant
  lever. The best wiring, an actively initialized residual bridge, reaches $96.2\%$, matching the far more elaborate three-expert
  AffordanceVLA ($95.8\%$) with under $1\%$ extra parameters. Two probes explain the mechanism: ground-truth affordances fed as an input
  hurt, and inference-time zeroing shows a lazy bridge acts only as a training-time regularizer while an active bridge becomes
  load-bearing.
\end{abstract}
\section{INTRODUCTION}

Vision-language-action (VLA) models have rapidly become the dominant recipe for language-conditioned robot manipulation. A large vision-language backbone, pretrained on internet-scale image-text data, is fine-tuned by imitation to map camera observations and a natural-language instruction to low-level actions \cite{rt2,openvla,octo,pi0,groot}. The recipe inherits the backbone's open-vocabulary perception and its language grounding, and recent flow-matching action experts \cite{pi0} have pushed success rates on standard benchmarks such as LIBERO \cite{libero} above the 90\% mark. Yet the supervision these models receive is curiously indirect. The action loss constrains only the \emph{trajectory}; nothing in the objective ever tells the model \emph{where} in the scene the interaction should happen. Which object the instruction refers to, which part of it to grasp, and where to place it must all be discovered implicitly, by correlating pixels with future gripper positions across the demonstration corpus.

The failure modes of imitation-trained VLAs mirror this gap. In our experiments, a strong 2B-parameter policy is nearly saturated on suites that test object diversity or spatial layout ($98{+}\%$), but loses most of its remaining errors in two places. The first is \emph{instruction disambiguation}: scenes containing several plausible targets where only the language picks out the right one. The second is \emph{long-horizon tasks}, where a small placement error early in the episode compounds. Both are precisely the situations in which an explicit, spatially grounded notion of ``the place to act next'' should help.

Affordances offer exactly that notion. The term, originating in ecological psychology \cite{gibson}, denotes the action possibilities a scene offers an agent; in robot learning it has been operationalized as dense maps of contact or interaction regions \cite{where2act,vrb,moka}. Affordance labels are attractive as auxiliary supervision for three reasons. First, in simulation they are \emph{free}: the physics engine knows every contact between gripper and object, so pixel-accurate interaction maps can be extracted from existing demonstrations without human labeling. Second, they are \emph{dense}: a per-pixel heatmap carries orders of magnitude more spatial signal per frame than a single action vector. Third, they are \emph{task-aligned}: predicting where to interact next requires resolving the instruction, binding it to an object, and localizing a part, which are the exact sub-skills whose failures dominate VLA error budgets.

It is therefore no surprise that several recent systems build affordances into VLAs \cite{deva,affordancevla}. What is surprising is how inconsistent the reported outcomes are, and how much architectural machinery is deployed to make the signal help. AffordanceVLA \cite{affordancevla}, the strongest published system, introduces a purpose-built three-expert mixture-of-transformers with a dedicated affordance-generation expert and a two-stage curriculum, and its own ablation shows that removing the second stage costs almost ten points. Our starting point was the opposite experience: attaching a straightforward affordance head to a VLA and co-training \emph{hurt}, collapsing instruction-following performance. The same auxiliary task, on the same benchmark, can evidently help or harm depending on integration details that papers rarely state, let alone ablate.

This paper treats those details as the object of study. Our method rests on three wiring decisions. The first is a \textbf{stop-gradient}: the affordance head reads intermediate backbone layers, but its loss never updates the backbone. This is not a stylistic choice. Ablating it flips the sign of the whole intervention, dropping goal-conditioned success from the $90$s to $62\%$ and making the head-equipped policy \emph{worse} than the headless base. With the backbone protected, the head's only channel of influence on the policy is a single learned \textbf{bridge} that re-injects an intermediate head feature into the visual tokens conditioning the action expert. This bridge is defined by two remaining bits of wiring, and we ablate both under a strictly identical data, budget, and seed protocol:

\begin{figure*}[t]
\centering
\includegraphics[width=0.95\textwidth]{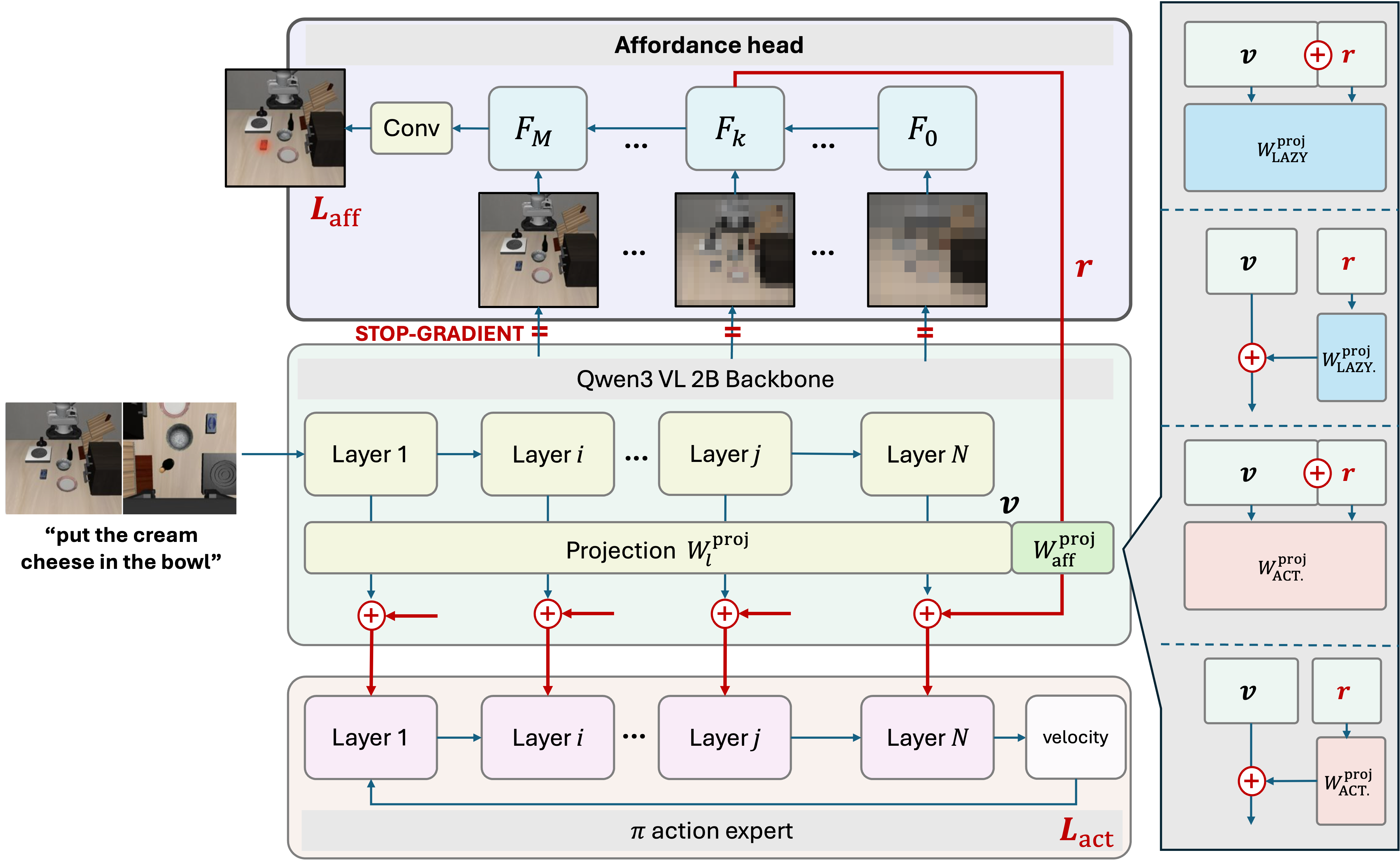}
\caption{\textbf{Left:} our recipe (\resactive{}). The backbone encodes the camera views and the instruction in a single forward pass; per-layer projections $W^{\mathrm{proj}}_l$ hand each layer's visual tokens $v$ to the corresponding layer of the flow-matching action expert. The affordance head reads uniformly spaced backbone layers through stop-gradient taps, reassembles earlier layers onto finer grids and later layers onto coarser ones, fuses them through a cascade of stages $F_0,\dots,F_M$, and decodes a per-view contact heatmap supervised by $\mathcal{L}_{\mathrm{aff}}$. The feature $r$ of an intermediate fusion stage $F_k$ is carried by the bridge $W^{\mathrm{proj}}_{\mathrm{aff}}$ (written $W_{\mathrm{aff}}$ in the text) back into the visual tokens at every conditioning layer. The expert is trained with the action objective $\mathcal{L}_{\mathrm{act}}$ in a single pass; at inference it integrates the learned flow over a few passes. \textbf{Right:} the four bridge wirings of our $2{\times}2$ ablation, top to bottom \catlazy{}, \reslazy{}, \catactive{}, and \resactive{} (ours). Blue denotes a zero-initialized (lazy) bridge and red a randomly initialized (active) one; a $\oplus$ joining the adjacent $v$ and $r$ boxes denotes concatenation before the shared projection, whereas a $\oplus$ on the token stream denotes elementwise addition of the projected $r$ onto the untouched $v$.}
\label{fig:arch}
\end{figure*}

\begin{itemize}
\item \textbf{Injection topology}: the head feature is either \emph{concatenated} with the visual tokens and linearly re-projected, which re-mixes (and can distort) the pretrained features, or added as a \emph{residual}, which preserves them exactly.
\item \textbf{Bridge initialization}: the injection weight starts at \emph{zero} (``lazy'': the policy begins identical to the base and must earn a dependence on the head) or at a standard \emph{random} initialization (``active'': the head's feature participates at full magnitude from the first step).
\end{itemize}

These bits swing the result by more than the presence of the head itself. Initialization is the dominant lever, worth $+2.0$ to $+2.7$ points on either topology; topology contributes a further half point, and only under active initialization. The best cell, residual injection with active initialization, lifts the average success on LIBERO from $93.1\%$ to $96.2\%$, with the gains concentrated exactly on the two failure modes above (Goal $+5.6$, Long $+6.2$).

\noindent\textbf{Contributions.}
(1)~A simple affordance-head recipe (stop-gradient + feature re-injection) that lifts a strong VLA from $93.1\%$ to $96.2\%$ average success on LIBERO, on par with the dedicated mixture-of-transformers AffordanceVLA ($95.8\%$) \cite{affordancevla} at a fraction of its architectural complexity.
(2)~A stop-gradient ablation showing gradient isolation is a \emph{precondition}: with affordance gradients flowing into the backbone, the head makes the policy \emph{worse} than the headless base ($85.5\%$ vs.\ $93.1\%$, with goal-conditioned success collapsing to $62\%$). The sign of the entire intervention flips on this single bit.
(3)~A same-budget $2{\times}2$ factorial ablation showing that, once the backbone is protected, \emph{initialization of the injection bridge, not its topology, is the dominant lever}. Active initialization adds $+2.0$/$+2.7$ points over lazy on the concatenation/residual topologies respectively, a two-way internal replication. Topology contributes at most $+0.5$, and only under active initialization.
(4)~Two mechanism probes. Ground-truth affordance maps supplied as an \emph{input} hurt ($-5.6$ points), showing the signal must be internalized by prediction rather than injected as data. Inference-time zeroing of the injected feature is a free diagnostic that separates two regimes. Under a lazy bridge the head is a pure training-time regularizer whose benefit survives ablation. Under an active bridge the affordance pathway becomes load-bearing, and it additionally lifts long-horizon success from $82.6\%$ to $91.4\%$.

\section{RELATED WORK}

\subsection{Vision-language-action models}
Casting manipulation as next-action prediction on top of a pretrained vision-language model was popularized by RT-2 \cite{rt2}, which showed that web-scale semantics transfer into control when actions are emitted as tokens. Subsequent open efforts scaled the recipe across embodiments and datasets: Octo \cite{octo} and OpenVLA \cite{openvla} established open-weight generalist policies, $\pi_0$ \cite{pi0} replaced autoregressive action decoding with a flow-matching action expert conditioned on backbone features, and GR00T~N1 \cite{groot} extended the approach toward humanoid platforms. A parallel line augments the visual side of VLAs with explicit spatial structure, e.g.\ 3D position encodings in SpatialVLA \cite{spatialvla} or intermediate visual sub-goals in CoT-VLA \cite{cotvla}. Our study is orthogonal to the backbone choice and concerns only how auxiliary dense supervision should be attached.

\subsection{The StarVLA platform and the QwenPI model}
\label{sec:qwenpi}
We conduct all experiments in StarVLA \cite{starvla}, an open-source, ``Lego-like'' codebase for VLA development in which every functional component (model, data pipeline, trainer, configuration, and benchmark evaluation) is factored into plug-and-play modules with high cohesion and low coupling. The platform lets a VLM backbone be paired with interchangeable action experts and evaluated under standardized pipelines across benchmarks including LIBERO, RoboCasa, and CALVIN. This modularity is not incidental to our study but instrumental. A controlled wiring ablation is only meaningful if inserting the head, swapping the bridge topology, or changing an initialization touches \emph{nothing else} in the stack. StarVLA's factoring makes each of our variants a small, isolated diff against an otherwise frozen training and evaluation pipeline.

Our base policy is StarVLA's \textbf{QwenPI} instantiation, which realizes the $\pi_0$ design \cite{pi0} on a modern open backbone. A Qwen3-VL-2B VLM \cite{qwenvl}, a natively multimodal model pairing a ViT vision tower with a language decoder, encodes the camera views and the instruction. Action generation is delegated to a \emph{layerwise flow-matching action expert}: a DiT that carries a noisy action chunk and denoises it under a conditional flow-matching objective, attending at each of its layers to the hidden states of the corresponding VLM layer. Compared with autoregressive action tokenization \cite{rt2,openvla}, this keeps actions continuous and chunked while letting the expert read the backbone at every depth, which is precisely the interface our affordance bridge later writes into.

\subsection{Affordances for manipulation}
The original notion of affordance, the action possibilities an environment offers \cite{gibson}, entered robot learning as the problem of predicting \emph{where} and \emph{how} an agent can interact. Where2Act \cite{where2act} learns per-pixel actionability scores for articulated objects from simulated interaction; VRB \cite{vrb} distills contact regions and post-contact trajectories from human videos; MOKA \cite{moka} elicits affordance points from a VLM zero-shot via visual prompting and executes them open-loop. These works treat the affordance predictor as the \emph{product}. Closer to us are systems that use affordances \emph{in service of} an end-to-end policy: DeVA \cite{deva} co-trains dense affordance and depth heads on a video-generation backbone for embodied control, and AffordanceVLA \cite{affordancevla} interleaves an affordance-generation expert with action generation inside a three-expert mixture-of-transformers, trained with a two-stage curriculum. Both report that affordance awareness improves manipulation, but both couple the signal to substantial bespoke architecture, which leaves open what the \emph{minimal} effective coupling is. Our result, a stock VLA plus one head and one linear bridge matching AffordanceVLA on LIBERO, suggests the architecture is not the essential ingredient; the wiring is.

\subsection{Auxiliary objectives, gradient interference, and bridges}
Auxiliary losses have long been used to shape representations, from the self-supervised auxiliary tasks of UNREAL in RL \cite{unreal} to depth and segmentation heads in perception stacks. The risk is equally well documented: auxiliary gradients can interfere destructively with the primary task \cite{pcgrad}. Our preliminary experiments reproduce exactly this failure, with the affordance loss collapsing instruction following when backpropagated into the backbone. Our stop-gradient design sidesteps interference structurally rather than algorithmically, at the price of raising a new question: if the auxiliary task cannot alter the backbone, how does it influence the policy at all? The answer runs through the injection bridge, which connects our work to a literature on \emph{how to attach new pathways to pretrained models}. Zero-initialized connections are the standard conservative choice: ControlNet's zero convolutions \cite{controlnet}, Flamingo's tanh-gated cross-attention \cite{flamingo}, and the zero-initialized up-projection of LoRA \cite{lora} all guarantee that a new branch starts as an exact identity and grows only as needed. Our central empirical finding is that this orthodoxy can silently fail in the auxiliary-head setting. When the primary task can already be solved from the existing features, the zero-initialized bridge sits at a \emph{lazy attractor}: it never grows, the injected feature stays runtime-inert, and the head's benefit reduces to a real but smaller implicit regularization effect. Random initialization is exactly what the adapter literature avoids, because it perturbs the pretrained computation. Yet it is what forces the policy's representation to form around the affordance pathway, and it is worth more than two points on LIBERO. This inverts the usual guidance and, to our knowledge, has not been isolated before in a VLA setting.

\section{METHOD}
\label{sec:method}

We describe our full method as instantiated in its best configuration (\resactive{}); Sec.~\ref{sec:wiring} defines the three ablated alternatives. Fig.~\ref{fig:arch} summarizes the data flow.

\subsection{Base policy and notation}
The base policy is the QwenPI model introduced in Sec.~\ref{sec:qwenpi}: the Qwen3-VL-2B backbone encodes two RGB views (a third-person and a wrist camera) and the instruction, and the layerwise flow-matching expert denoises action chunks. The backbone has $N$ layers; we write $H_l$ for the hidden states of backbone layer $l$ and $v_l$ for the visual tokens the expert consumes at conditioning layer $l$. Writing $x_1$ for the demonstrated action chunk and $x_0$ for Gaussian noise, the expert $v_\theta$ is trained with the conditional flow-matching objective \cite{pi0}
\begin{equation}
\mathcal{L}_{\mathrm{act}}
= \mathbb{E}_{t,\,x_0}\,\big\|\, v_\theta\big(x_t,\, t,\, \{v_l\}\big) - (x_1 - x_0) \,\big\|^2,
\label{eq:fm}
\end{equation}
where $x_t = (1{-}t)\,x_0 + t\,x_1$ is the linearly interpolated noisy chunk. The base is trained with $\mathcal{L}_{\mathrm{act}}$ only.

\subsection{Affordance supervision from demonstration replay}
\label{sec:labels}
Our affordance target answers one question per frame: \emph{where will the gripper next make task-relevant contact?} We extract this from the existing LIBERO demonstrations, with no human labeling, by exploiting privileged simulator state. LIBERO tasks run in a MuJoCo physics engine \cite{mujoco} via the robosuite framework \cite{robosuite}; each demonstration is replayed step by step, and at every timestep the engine's contact solver exposes the exact set of geometry pairs in contact. Scanning \emph{forward} along the replay from frame $t$, the first object the gripper contacts defines the frame's interaction target. Its 3D contact location is projected through each camera's calibrated intrinsics and extrinsics to a pixel coordinate $u^{*}$, and the label is an isotropic Gaussian of fixed scale $\sigma$ centered there, one map per camera view:
\begin{equation}
A(u) \;=\; \exp\!\Big(\!-\tfrac{\|u - u^{*}\|^2}{2\sigma^2}\Big).
\label{eq:gt}
\end{equation}

Two properties of this construction matter. First, the label is a \emph{look-ahead} signal: within a task stage it is constant (the whole approach phase points at the same grasp location) and it switches discretely at stage boundaries (the moment the mug is grasped, the label jumps to the target plate). The head is therefore supervised to predict not the present but the immediate motor future, the same latent decision the policy must make. Second, because the Gaussian template is fixed, the per-frame label mass is constant, so the prediction task never degenerates into predicting label \emph{intensity}; only \emph{location} carries information.

\subsection{Affordance head: multi-scale readout of a columnar backbone}
\label{sec:head}
\textbf{Where the head reads from.} The head taps the backbone's hidden states at a small set of uniformly spaced layers. The motivation is the well-established semantic stratification of deep networks: early transformer layers carry local, high-frequency appearance information, while late layers carry increasingly global, language-conditioned semantics. Affordance prediction needs \emph{both ends} of this spectrum simultaneously. Selecting the correct object among distractors is a late-layer, instruction-conditioned decision, but placing the peak on the mug's \emph{handle} to pixel accuracy is an early-layer, appearance-driven one. A single tap would force a choice between the two; uniformly spaced taps hand the head the full hierarchy and let it learn the mixture.

\textbf{Why a pyramid.} A vision transformer is \emph{columnar}: every layer represents the image at the same, coarse token resolution, unlike convolutional hierarchies whose resolution decreases with depth \cite{vit}. Dense prediction from such a backbone therefore faces the classic tension between recognition and localization. Semantic strength lives in coarse features, while spatial precision requires fine ones. The canonical resolution of this tension is the top-down pyramid with lateral connections, introduced by FPN for object detection \cite{fpn} and, in encoder-decoder form, by U-Net for segmentation \cite{unet}. In such a pyramid, coarse and semantically strong features are progressively upsampled and \emph{fused at each scale} with finer ones, so every output location is informed by both global context and local evidence. DPT \cite{dpt} showed how to manufacture the missing multi-scale structure from a columnar ViT by \emph{reassembling} different layers' tokens at different synthetic resolutions before running such a pyramid. We adopt this principle for the same first-principles reason it was invented: our output must be sharp at pixel level but selected by global, instruction-level semantics. Concretely, each tapped layer is reassembled onto its own grid, earlier layers onto finer grids to preserve their high-frequency content and later layers onto coarser grids matching their semantic granularity. The grids are fused top-down with lateral connections through a cascade of fusion stages $F_0,\dots,F_M$ (coarse to fine) and decoded to the per-view contact heatmap, trained with binary cross-entropy $\mathcal{L}_{\mathrm{aff}}$.

\textbf{Instruction conditioning.} One subtlety of reading a causal multimodal backbone is that the image tokens precede the instruction in the sequence, so under causal attention their hidden states cannot attend to the text that follows. The head therefore receives the instruction through an explicit second route: the instruction token embeddings are pooled and applied as a FiLM modulation \cite{film} to the readouts and to every fusion stage. The head is thus instruction-conditioned by construction, which is what allows it to place its peak on the commanded object even when an identical distractor is present. It also means the injected feature $r$ carries an instruction-resolved spatial decision rather than a purely visual saliency map.

\textbf{Stop-gradient.} Collecting the pieces, the head $h_\phi$ maps the tapped hidden states and the pooled instruction embedding $e_{\mathrm{txt}}$ to a predicted map, and is trained against the label of Eq.~\eqref{eq:gt} with binary cross-entropy:
\begin{equation}
\hat{A} \;=\; h_\phi\big(\{\,\mathrm{sg}[H_l]\,\}_{l\in\mathcal{T}},\; e_{\mathrm{txt}}\big),
\qquad
\mathcal{L}_{\mathrm{aff}} \;=\; \mathrm{BCE}\big(\hat{A},\, A\big),
\label{eq:head}
\end{equation}
where $\mathcal{T}$ is the set of tapped layers and $\mathrm{sg}[\cdot]$ is the stop-gradient operator: $\mathcal{L}_{\mathrm{aff}}$ updates \emph{only} head parameters $\phi$, never the backbone. The rationale is asymmetric protection. The backbone's language grounding is the scarcest resource in the system; it was bought with internet-scale pretraining and is easily corrupted. In preliminary experiments, letting affordance gradients into the backbone collapsed goal-conditioned success from $94\%$ to below $32\%$. The head, in contrast, is cheap and can adapt to the backbone rather than vice versa. Note that the stop-gradient is not a freeze. Every parameter of the backbone and the action expert remains trainable end to end under the action loss; we only deny the auxiliary loss a write path into the backbone. The recipe therefore needs no frozen modules, no LoRA-style protection, and no staged curriculum: it is plain full fine-tuning, single-stage, under the combined objective
\begin{equation}
\mathcal{L} \;=\; \mathcal{L}_{\mathrm{act}} + \lambda\,\mathcal{L}_{\mathrm{aff}},
\label{eq:total}
\end{equation}
and we found performance insensitive to $\lambda$.

\subsection{The affordance bridge}
\label{sec:wiring}
The stop-gradient leaves the head an isolated observer: accurate, but by default mute with respect to the policy. Everything the head contributes to control therefore flows through a single learned \textbf{bridge}, and the rest of this section develops the bridge from the inside out: which head feature it carries, and the small design space of ways to write that feature into the policy.

The carried feature is not the final heatmap. The heatmap is a one-channel bottleneck: it reports \emph{where} but discards the head's intermediate computation about \emph{what} and \emph{why}. We instead take the feature $r$ of an intermediate fusion stage $F_k$, chosen because its spatial grid coincides with the backbone's native visual-token grid. Injection is then token-aligned: each visual token receives the head feature for precisely its own spatial cell. At the same time, $F_k$ sits late enough in the cascade that coarse semantics and fine localization have already been married. This feature is written into the visual tokens at \emph{every} conditioning layer of the action expert, so the affordance evidence accompanies the visual stream at whatever depth the expert consults it. Our default writes it as a residual:
\begin{equation}
v_l' \;=\; v_l + W_{\mathrm{aff}}\, r \qquad \text{(residual)},
\label{eq:res}
\end{equation}
on the standard grounds of residual learning \cite{resnet}: the pretrained visual features, the base policy's entire working capital, pass through unmodified, and the head contributes a purely additive refinement. The natural alternative concatenates and re-projects,
\begin{equation}
v_l' \;=\; W_{\mathrm{aff}}\,[\, v_l \,\|\, r\,] \qquad \text{(concatenation)},
\label{eq:cat}
\end{equation}
which is more expressive but forces even the original features through a learned re-mixing; Sec.~\ref{sec:mainresult} shows this measurably erodes fine-grained placement.

The bridge has one further degree of freedom, and it turns out to be the consequential one: how $W_{\mathrm{aff}}$ is initialized. We use a standard Kaiming random scheme (\emph{active}) rather than zero (\emph{lazy}), contradicting the prevailing adapter-design orthodoxy \cite{controlnet,flamingo,lora}. The zero-initialized bridge looks safer, since the policy starts exactly at the base model. But it must earn its growth from the gradient of the action loss, and that gradient is starved by construction: the head offers information the action expert can, with effort, extract from $v_l$ anyway. The marginal value of the bridge at $W_{\mathrm{aff}}{=}0$ is therefore near zero, and the optimizer never leaves the attractor. Empirically, lazily initialized bridges converge to a few percent of the magnitude of the identity path and are runtime-inert; zeroing them at inference changes nothing (Sec.~\ref{sec:zeroing}). Active initialization removes the chicken-and-egg problem by fiat. The feature $r$ is present at full magnitude from step one, the expert's representation forms around it, and the pathway becomes genuinely load-bearing, at the cost of a violent start: the action loss opens an order of magnitude higher before recovering within a few hundred steps.

Crossing the two topologies with the two initializations yields the four wirings we evaluate (Fig.~\ref{fig:arch}, right): \catlazy{}, \reslazy{}, \catactive{}, and \resactive{}, of which \resactive{} is our method and the other three are ablations.

\subsection{What can the head even contribute?}
The affordance target is a deterministic function of the observation and instruction, and the head reads the same backbone features the action expert already sees; information-theoretically, $r$ can offer the policy \emph{no new information}, only a more explicit format. Any benefit must therefore come from (a) the optimization dynamics induced by the head's presence, or (b) the explicit format easing the action expert's job at run time. The two initializations of Sec.~\ref{sec:wiring} turn out to select between exactly these mechanisms, and the experiments below separate them.

\section{EXPERIMENTS}

\subsection{Setup}
All models are fine-tuned on the LIBERO demonstration corpus (all four suites jointly) from the same Qwen3-VL-2B initialization, with an identical budget: global batch 128, 20k steps ($\approx$9.4 epochs), identical hyperparameters and seed. The \emph{only} difference between the five runs is the head wiring. Evaluation is 50 trials per task (500 per suite) on LIBERO-Spatial, -Object, -Goal, and -Long (LIBERO-10).

\begin{table}[t]
\caption{LIBERO success rates (\%), 50 trials/task. All runs share data, budget, and seed; only the head wiring differs. \textbf{Bold} = best per column. $^\dagger$Measured in a preliminary configuration of the same pipeline.}
\label{tab:main}
\centering
\setlength{\tabcolsep}{4.5pt}
\begin{tabular}{lccccc}
\toprule
Model & Object & Spatial & Goal & Long & Avg \\
\midrule
Base (no head) & 98.0 & 98.6 & 90.6 & 85.2 & 93.1 \\
Head w/o stop-grad$^\dagger$ & 100.0 & 92.0 & 62.0 & 88.0 & 85.5 \\
\midrule
\catlazy{} & \textbf{99.4} & 98.2 & 94.8 & 82.6 & 93.8 \\
\reslazy{} & 98.8 & 96.8 & 93.4 & 85.2 & 93.6 \\
\catactive{} & 98.0 & \textbf{99.6} & 96.0 & 89.4 & 95.8 \\
\resactive{} & 98.8 & 98.4 & \textbf{96.2} & \textbf{91.4} & \textbf{96.2} \\
\bottomrule
\end{tabular}
\end{table}

\begin{figure*}[!t]
\centering
\includegraphics[width=0.92\textwidth]{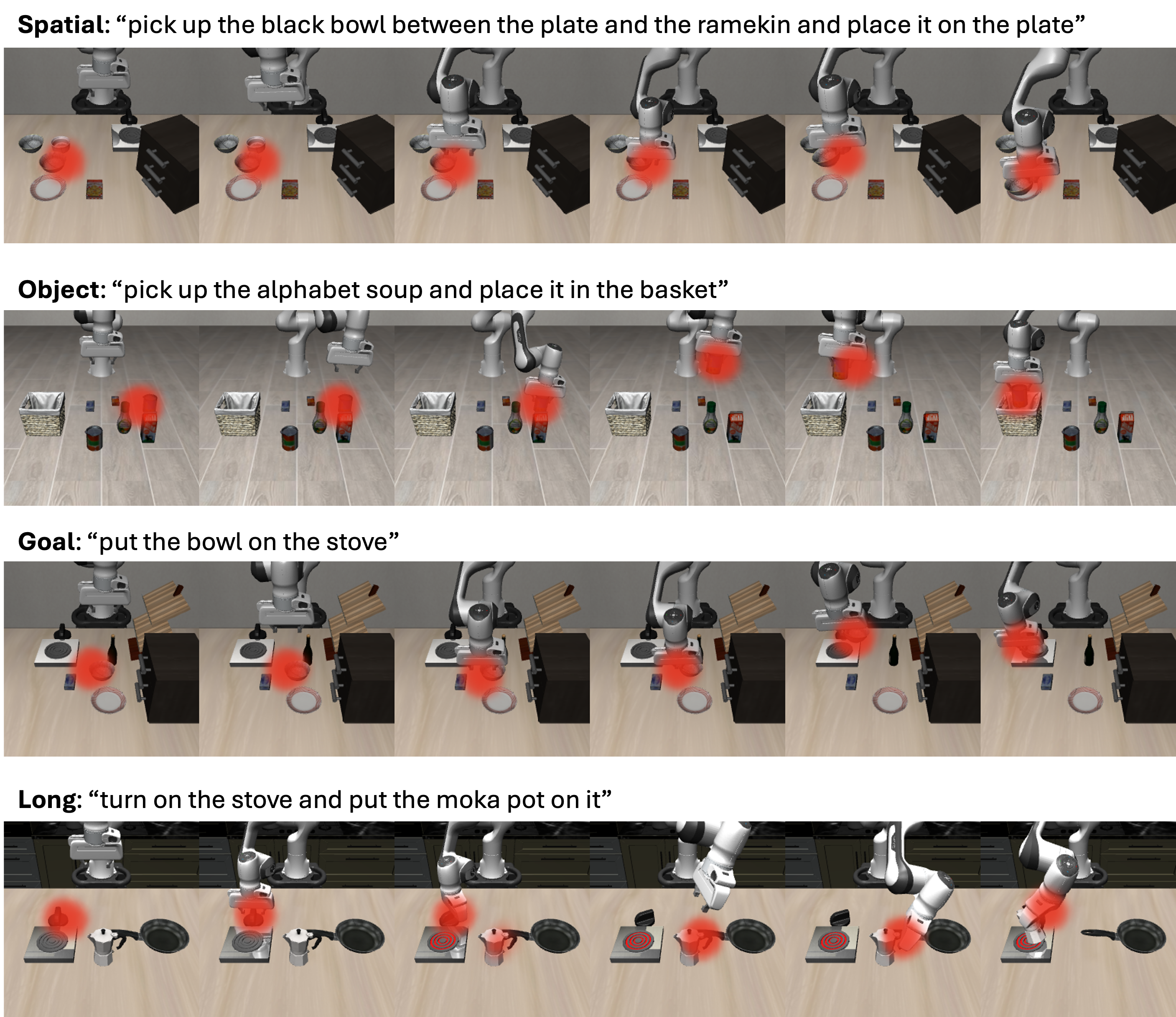}
\caption{\resactive{} rollouts with the affordance head's own predictions overlaid in red. One task per suite, six evenly spaced frames per episode; all four episodes succeed. The predicted contact region locks onto the instruction-designated object among distractors and switches discretely at stage boundaries, e.g.\ from the stove knob to the moka pot to the burner in the long-horizon task.}
\label{fig:rollouts}
\end{figure*}

\begin{figure*}[!t]
\centering
\includegraphics[width=0.95\textwidth]{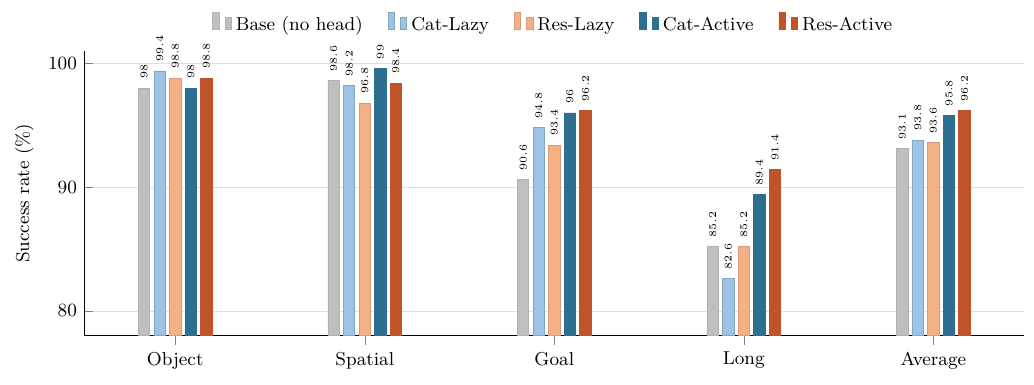}
\caption{Per-suite success rates of the headless base and the four bridge wirings (50 trials/task). Color encodes topology (blue: concatenation, orange: residual) and saturation encodes initialization (light: lazy, dark: active). The two active wirings dominate on Goal and Long, the suites that stress instruction disambiguation and long-horizon precision, while all five models are near saturation on Object and Spatial.}
\label{fig:bars}
\end{figure*}

\subsection{Main result: the head helps, and wiring decides how much}
\label{sec:mainresult}
Table~\ref{tab:main} and Fig.~\ref{fig:bars} show the full $2{\times}2$ plus the headless base. Every wired variant matches or beats the base on average, and the best cell, \resactive{}, improves the average from $93.1\%$ to $96.2\%$, with the gains exactly where a policy needs help: Goal $+5.6$ and Long $+6.2$, at parity on the near-saturated Object/Spatial suites. Reading the table factorially:

\textbf{Initialization is the dominant lever.} Active vs.\ lazy is worth $+2.0$ points on the concatenation topology ($93.8{\rightarrow}95.8$) and $+2.7$ on the residual topology ($93.6{\rightarrow}96.2$): the same direction and magnitude on both rows, a two-way replication. Inspecting checkpoints explains why. With lazy initialization the bridge weight converges to only a few percent of the magnitude of the identity path, i.e.\ the optimizer never grows a real dependence on $r$; with active initialization the bridge retains full magnitude and the action expert's representation forms around it.

\textbf{Topology is a secondary lever, visible only under active init.} Under lazy initialization the two topologies tie ($93.8$ vs.\ $93.6$). Under active initialization the residual form adds $+0.5$ over concatenation, with the gain concentrated on LIBERO-Long ($89.4{\rightarrow}91.4$). The Long-suite column tells the cleanest story: $82.6$ (\catlazy) $\rightarrow 85.2$ (\reslazy) $\rightarrow 89.4$ (\catactive) $\rightarrow 91.4$ (\resactive). Concatenation's re-mixing of visual tokens measurably harms fine-grained, long-horizon placement: under lazy initialization the bridge is inert, so re-mixing is the only active difference, and it costs $2.6$ points against the base. The residual's identity path recovers exactly that loss. Active initialization then adds its larger, orthogonal gain on top.

\begin{figure}[t]
\centering
\includegraphics[width=\linewidth]{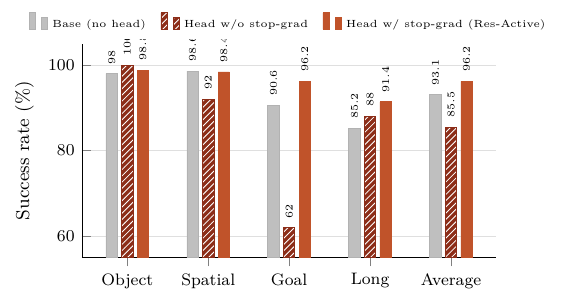}
\caption{The stop-gradient is necessary. The same head helps or hurts depending on this single bit: with the gradient blocked (\resactive{}) the head lifts the average to $96.2$, while letting the affordance gradient reach the backbone drops it to $85.5$, a $10.7$-point swing. The gradient-exposed variant even falls $7.6$ points below the headless base, and the damage is selective: Goal, the suite that leans hardest on instruction grounding, collapses from the $90$s to $62.0$.}
\label{fig:stopgrad}
\end{figure}

\subsection{Stop-gradient is the precondition}
Fig.~\ref{fig:stopgrad} makes the case with a three-way comparison: the headless base, the head with its gradient allowed into the backbone, and the same head behind a stop-gradient (\resactive{}). The first comparison is against our best model. Blocking the gradient is worth $10.7$ points on average ($85.5$ vs.\ $96.2$), so the stop-gradient is not an optimization detail but the difference between the head helping and the head failing. The second comparison is against the base, and it is the more damning one. The gradient-exposed variant lands $7.6$ points \emph{below} a model with no head at all: the auxiliary gradient does not merely waste the head, it actively degrades the backbone's own reasoning. The damage is selective and diagnostic. Object-level success is untouched (a perfect $100$), since recognizing and reaching a unique object needs little language. Goal-conditioned success instead collapses to $62\%$. The auxiliary gradient specifically erodes the backbone's \emph{instruction grounding}, the capability bought with vision-language pretraining and consumed most heavily by disambiguation. We suspect this failure mode accounts for much of the mixed evidence on auxiliary heads in the literature. All results in the rest of the paper therefore keep the stop-gradient in place; it is the precondition that makes the remaining $2{\times}2$ worth studying.

\subsection{Comparison with AffordanceVLA}
Table~\ref{tab:avla} compares our best variant with AffordanceVLA \cite{affordancevla}, the closest prior work. The two reach essentially the same performance, but by very different means. AffordanceVLA introduces a purpose-built three-expert Mixture-of-Transformers with a dedicated affordance-generation expert and a two-stage training curriculum, and its own ablation shows the design is fragile, dropping to $86.2\%$ without the second stage. Our result comes from a stock VLA plus one DPT-style head and one linear bridge, which together add under $1\%$ to the policy's parameter count. Training is a single stage with no frozen modules, and the inference-time overhead of running the head is negligible. The entire ``architecture'' of our method is two wiring decisions. Explicit affordance grounding, in other words, does not require an affordance-specialized architecture. It requires getting the wiring right.

\begin{table}[t]
\caption{Comparison with AffordanceVLA on LIBERO (\%). Their numbers as reported in \cite{affordancevla}.}
\label{tab:avla}
\centering
\setlength{\tabcolsep}{3.5pt}
\begin{tabular}{lccccc}
\toprule
Method & Object & Spatial & Goal & Long & Avg \\
\midrule
AffordanceVLA (w/o stage II) & 91.7 & 88.5 & 91.3 & 73.3 & 86.2 \\
AffordanceVLA (full) & 98.4 & \textbf{98.6} & \textbf{96.2} & 89.8 & 95.8 \\
\resactive{} (ours) & \textbf{98.8} & 98.4 & \textbf{96.2} & \textbf{91.4} & \textbf{96.2} \\
\bottomrule
\end{tabular}
\end{table}

\subsection{Probe 1: ground-truth affordances as input are net harmful}
Is the affordance signal itself valuable enough that the policy should simply be \emph{given} it? To test this, we trained a variant that receives the ground-truth affordance heatmap as a third input image, with an oracle rendering the map online at evaluation (in a preliminary configuration of the same pipeline, against its own matched base at $94.0\%$ average). Despite privileged information, this variant reaches only $88.3\%$, a net $-5.6$. It is roughly neutral on Object and Spatial but loses $5.5$ on Goal and $19$ on Long: it specifically damages instruction disambiguation and multi-stage tasks. Externally supplied answers create an input shortcut that competes with the instruction pathway; the signal must be \emph{internalized} by prediction, not injected as data.

\subsection{Probe 2: inference-time zeroing separates two regimes}
\label{sec:zeroing}
We replace the injected feature with zeros at inference; the head still runs, only the bridge is silenced. For \catlazy{}, nothing changes (Goal $94.8{\rightarrow}95.6$, Long $82.6{\rightarrow}83.2$, within noise). This is consistent with the small bridge magnitude: the run-time contribution of $r$ is negligible, yet the variant still beats the base on Goal by $+4.2$. The head's entire benefit in this regime is a \emph{training-time regularization} effect. Its presence alters the optimization trajectory of the action expert, not the test-time computation. In contrast, in preliminary experiments with an actively initialized bridge, the same ablation collapsed Goal success by $28$ points: an active pathway becomes load-bearing. The zeroing probe is free to run and we recommend it as a standard diagnostic for auxiliary-task pathways in VLAs.

\subsection{Head quality is not the bottleneck}
Across variants the head itself trains to high fidelity: its predicted contact peaks land within a few pixels of ground truth on held-out frames and correctly disambiguate even duplicate objects. Rollout inspection of the remaining Long-suite failures shows correct target selection with second-stage precision errors (e.g., placing onto an occupied burner). The open problem is thus not \emph{predicting} affordances but \emph{transducing} them into finer control.

\section{LIMITATIONS AND CONCLUSION}

Our study is limited to one simulated benchmark, one backbone scale, and one seed per cell, partially mitigated by the two-way replication of the initialization effect and a point-identical retest of the Long suite. Affordance labels also come from privileged simulator state. Within this scope the conclusions are sharp. First, a stop-gradient affordance head with feature re-injection reliably helps. Both actively initialized wirings beat the headless base, \catactive{} by $+2.7$ points on average and \resactive{} by $+3.1$, with the gains concentrated on instruction disambiguation and long-horizon precision. Second, the two wiring choices help for separable reasons. Active initialization is the main lever on both topologies: it forces the action expert to build its representation around the affordance pathway instead of leaving the bridge at its lazy attractor. The residual topology adds its own margin on top, chiefly on the Long suite, because its identity path hands the pretrained visual features to the expert untouched, while concatenation re-mixes and slightly degrades them. \resactive{} combines both advantages and is the best model in every comparison we ran, matching the purpose-built AffordanceVLA architecture with one head and one linear bridge. More broadly, the wiring bits that most papers leave unstated, bridge initialization and injection topology, swing the result by more than the presence of the head itself. We hope the $2{\times}2$ protocol and the zeroing probe become standard reporting practice for auxiliary heads in VLAs.


\end{document}